\documentclass[10pt,twocolumn,letterpaper]{article}

\usepackage{cvpr}
\usepackage{times}
\usepackage{epsfig}
\usepackage{graphicx}
\usepackage{amsmath}
\usepackage{amssymb}
\usepackage{booktabs}
\usepackage{multirow}
\usepackage{url}
\usepackage{subfigure}
\usepackage{color}
\usepackage{makecell}
\usepackage{authblk}

\newcommand{\RM}[1]{\textcolor{black}{#1}}

\usepackage{array}
\newcommand{\PreserveBackslash}[1]{\let\temp=\\#1\let\\=\temp}
\newcolumntype{C}[1]{>{\PreserveBackslash\centering}p{#1}}

\usepackage[pagebackref=true,breaklinks=true,letterpaper=true,colorlinks,bookmarks=false]{hyperref}

\cvprfinalcopy 
\def\cvprPaperID{2291} 

\ifcvprfinal\fi

\usepackage{subfiles}
\usepackage{algorithmicx,algorithm}
\usepackage{algpseudocode}
\floatname{algorithm}{\scriptsize{Algorithm}}
\usepackage{authblk}

\begin{document}

\title{P2B: Point-to-Box Network for 3D Object Tracking in Point Clouds}

\author{
Haozhe Qi,~Chen Feng,~Zhiguo Cao\thanks{Zhiguo Cao is corresponding author (zgcao@hust.edu.cn).}, ~Feng Zhao, and~Yang Xiao\\
\fontsize{11.1pt}{\baselineskip}\selectfont National Key Laboratory of Science and Technology on Multi-Spectral Information Processing, School of Artificial Intelligence and Automation, Huazhong University of Science and Technology, Wuhan 430074, China\\
\tt\small{{qihaozhe, chen$\_$feng, zgcao}@hust.edu.cn, fzhao@alumni.hust.edu.cn, Yang$\_$Xiao@hust.edu.cn}
}

\maketitle
\thispagestyle{empty}

\begin{abstract}
Towards 3D object tracking in point clouds, a novel point-to-box network termed P2B is proposed in an end-to-end learning manner. Our main idea is to first localize potential target centers in 3D search area embedded with target information. Then point-driven 3D target proposal and verification are executed jointly. In this way, the time-consuming 3D exhaustive search can be avoided. Specifically, we first sample seeds from the point clouds in template and search area respectively. Then, we execute permutation-invariant feature augmentation to embed target clues from template into search area seeds and represent them with target-specific features. Consequently, the augmented search area seeds regress the potential target centers via Hough voting. The centers are further strengthened with seed-wise targetness scores.  Finally, each center clusters its neighbors to leverage the ensemble power for joint 3D target proposal and verification. We apply PointNet++ as our backbone and experiments on KITTI tracking dataset demonstrate P2B's superiority ($\sim$10\%'s improvement over state-of-the-art). Note that P2B can run with 40FPS on a single NVIDIA 1080Ti GPU. Our code and model are available at \url{https://github.com/HaozheQi/P2B}.

\end{abstract}

\section{Introduction}
\label{sec:introduction}
\begin{figure}
\centering
\includegraphics[width=8.2cm,height=5.1cm]{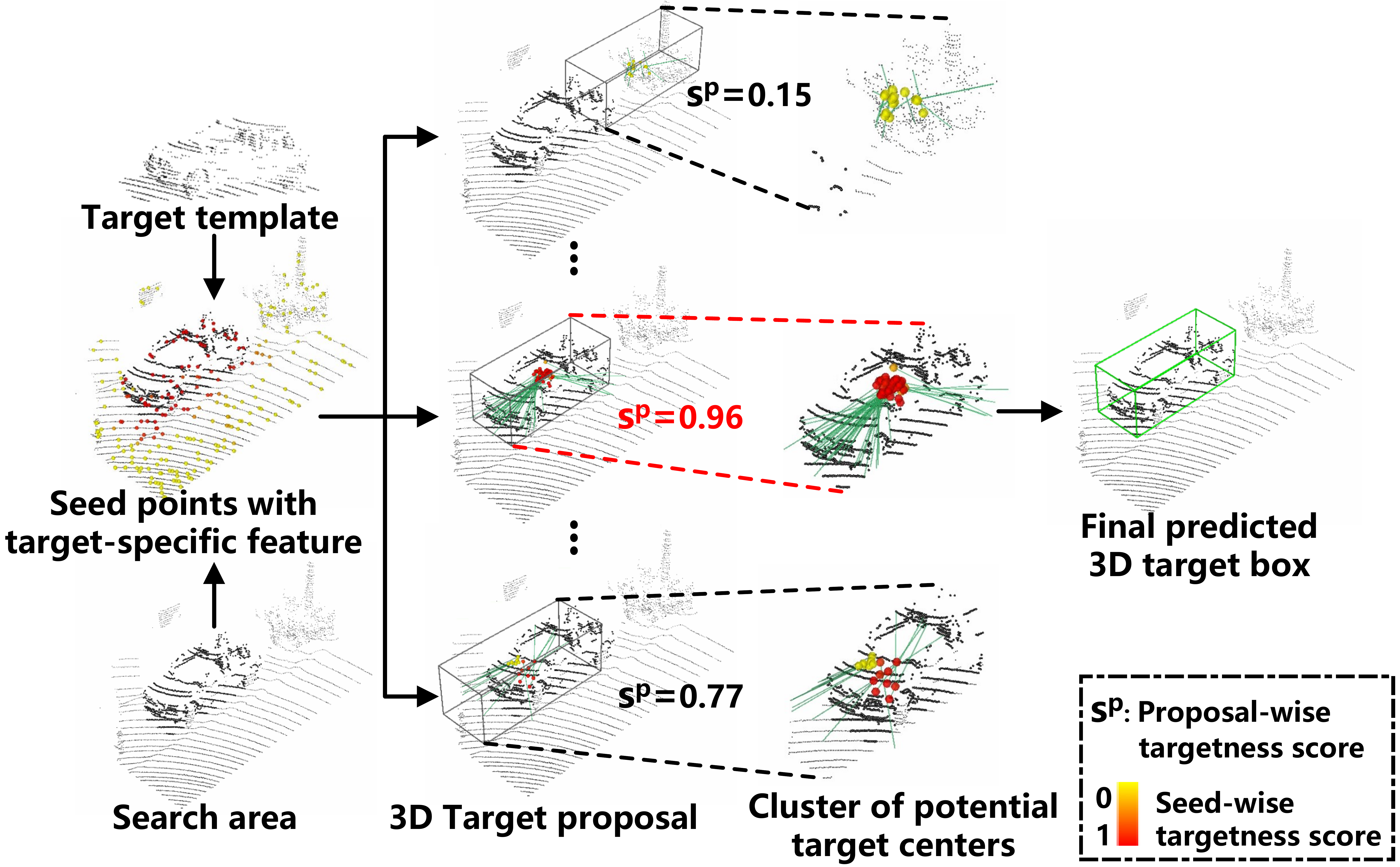}
\caption{\RM{\textbf{Exemplified illustration to show how P2B works}, from seeds sampling to 3D target proposal and verification.}}
\label{fig:fig1}
\vspace{-0.05cm}
\end{figure}

3D object tracking in point clouds is essential for applications in autonomous driving and robotics vision \cite{luo2018fast,machida2012human,comport2004robust}. However, point clouds' sparsity and disorder imposes great challenges on this task, and leads to the fact that, well-established 2D object tracking approaches (\textit{e.g.}, Siamese network \cite{siamFC}) cannot be directly applied. Most existing 3D object tracking methods~\cite{asvadi3d, Bibi3D, context, RGBDtracker, kartobject} inherit 2D's experience and rely heavily on RGB-D information. But they may fail when RGB visual information is degraded with illuminational change or even inaccessible. We hence focus on 3D object tracking using only point clouds.
\begin{figure*}[t]
	\centering
	\includegraphics[width=16cm]{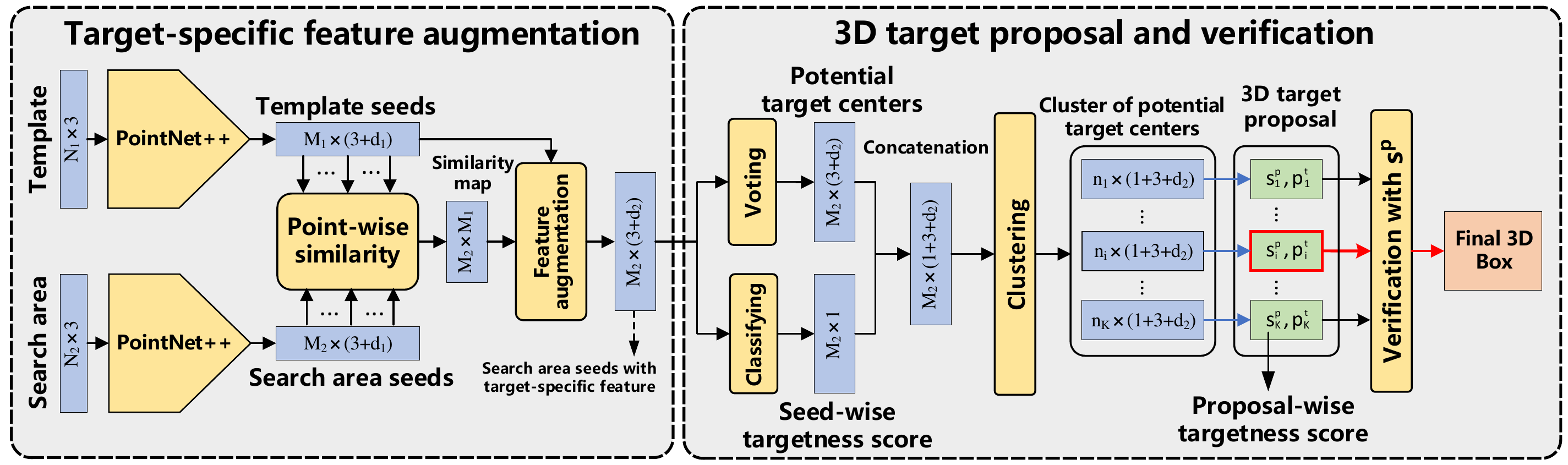}
	\caption{\textbf{The main pipeline of P2B}. P2B has two parts: 1) target-specific feature augmentation, 2) 3D target proposal and verification. The backbone applies modified PointNet++. 1) enriches search area seeds with target clue from template. With the augmented seeds, 2) regresses potential target centers and evaluates seed-wise targetness for joint target proposal and verification.}
	\label{fig:P2Bpipline}
\end{figure*}
The first pioneer effort on this topic appears in~\cite{leveraging}. It mainly executes 3D template matching using Kalman filtering \cite{gordon2004beyond} to generate bunches of 3D target proposals. Meanwhile, it uses shape completion to regularize feature learning on point set. Nevertheless, it tends to suffer from four main defects: 1) its tracking network cannot be end-to-end trained; 2) 3D search with Kalman filtering consumes much time; 3) each target proposal is represented with only one-dimensional global feature, which may lose fine local geometric information; 4) shape completion network brings strong class prior which weakens generality.

Towards the above concerns, we propose a novel point-to-box network termed P2B for 3D object tracking which can be end-to-end trained. Differing from the intuitive 3D search with box in~\cite{leveraging}, we turn to \emph{addressing 3D object tracking by first localizing potential target centers and then executing point-driven target proposal and verification jointly}. Our intuition lies in two folders. First, the point-wise tracking paradigm may help better exploit 3D local geometric information to characterize target in point clouds. Secondly, formulating 3D object tracking task in an end-to-end manner is of stronger ability to fit target's 3D appearance variation during tracking.

\RM{We exemplify how P2B works in Fig.~\ref{fig:fig1}}. We first feed template and search area into backbone respectively and obtain their seeds. The search area seeds will consequently predict potential target centers for joint target proposal and verification. Then the search area seeds are augmented with target-specific features, yielding three main components: 1) their 3D position coordinates to retain spatial geometric information, 2) their point-wise similarity with template seeds to mine resembling patterns and reveal the local tracking clue, and 3) encoded global feature of target from template. This augmentation is invariant to seeds' permutation and yields consistent target-specific features. After that, the augmented seeds are projected to the potential target centers via Hough voting~\cite{votenet}. Meanwhile, each seed is assessed with its targetness to regularize earlier feature learning; the result targetness score further strengthens its predicted target center's representation. Finally, each potential target center clusters the neighbors to leverage the ensemble power for joint target proposal and verification.

Experiments on KITTI tracking dataset \cite{KITTI} demonstrate that, P2B significantly outperforms the state-of-the-art method~\cite{leveraging} by large a margin ($\sim$10\% on both Success and Precision). Note that P2B can run with about 40FPS on a single NVIDIA 1080Ti GPU.

Overall, the main contributions of this paper include

$\bullet$ P2B: a novel point-to-box network for 3D object tracking in point clouds, which can be end-to-end trained;

$\bullet$ Target-specific feature augmentation to include global and local 3D visual clues for 3D object tracking;

$\bullet$ Integration of 3D target proposal and verification.

\section{Related Works}

We briefly introduce the works most related to our P2B: 3D object tracking, 2D Siamese tracking, deep learning on point set, target proposal and Hough voting.

\textbf{3D object tracking.} To the best of our knowledge, 3D object tracking using only point clouds has seldom been studied before the recent pioneer attempt~\cite{leveraging}. Earlier related tracking methods~\cite{context, RGBDtracker, kartobject,pieropan2015robust,asvadi3d,Bibi3D} generally resort to RGB-D information. Though with the paid efforts from different theoretical aspects, they may suffer from two main defects: 1) they rely on RGB visual clue and may fail if it is degraded or even inaccessible. This limits some real applications; 2) they have no networks designed for 3D tracking, which may limit the representative power. Besides, some of them~\cite{context, RGBDtracker, kartobject} focus on generating 2D boxes. The above concerns are addressed in~\cite{leveraging}. Leveraging deep learning on point set and 3D target proposal, it achieves the state-of-the-art result on 3D object tracking using only point clouds. However, it still suffers from some drawbacks as in Sec.~\ref{sec:introduction}, which motivates our research.

\textbf{2D Siamese tracking.} Numerous state-of-the-art 2D tracking methods~\cite{tao2016siamese, siamFC, wang2017dcfnet, held2016learning, zhu2018distractor, wang2018learning, li2019siamrpn, fan2019siamese, zhang2019deeper, wang2019fast, siamRPN} are built upon Siamese network. Generally, Siamese network has two branches for template and search area with shared weights to measure their similarity in an implicitly embedded space.
Recently, \cite{siamRPN} unites region proposal network and Siamese network to boost performance. Hence, time-consuming multi-scale search and online fine-tuning are both avoided. Afterwards, many efforts~\cite{zhu2018distractor,li2019siamrpn, zhang2019deeper,wang2019fast,fan2019siamese} follow this paradigm. However, the above methods are all driven by 2D CNN which is inapplicable to point clouds. We hence aim to extend the Siamese tracking paradigm to 3D object tracking with effective 3D target proposal.

\textbf{Deep learning on point set.} Recently, deep learning on point set draws increasing research interests~\cite{pointnet,pointnet++}. To address point clouds' disorder, sparsity and rotation variance, the paid efforts have facilitated the research in 3D object recognition~\cite{Escape,li2018pointcnn}, 3D object detection~\cite{votenet,frustum,PointRCNN,Yang_2019_ICCV_STD}, 3D pose estimation~\cite{Point_To_Pose,HandPointNet,chen2018shpr}, and 3D object tracking~\cite{leveraging}.
However, the 3D tracking network in~\cite{leveraging} cannot execute end-to-end 3D target proposal and verification jointly, which constitutes P2B's focus.


\textbf{Target proposal.} In 2D tracking tasks, many tracking-by-detection methods \cite{Zhu2016BeyondLS,wang2019GANTrack,huang2019GlobalTrack} exploit the target clue contained in template to obtain high-quality target-specific proposals. They operate on (2D) area-based pixels with either edge features \cite{Zhu2016BeyondLS}, region-proposal network \cite{wang2019GANTrack} or attention map \cite{huang2019GlobalTrack} in a target-aware manner. Comparatively, P2B regards each point as a regressor towards potential target center which directly relates to 3D target proposal.

\textbf{Hough voting.} The seminal work of Hough voting~\cite{houghvoting} proposes a highly flexible learned representation for object shape, which can combine the information observed on different training examples in a probabilistic extension of the Generalized Hough Transform~\cite{houghtransform}. Recently, \cite{votenet} embeds Hough voting into an end-to-end trainable deep network for 3D object detection in point cloud, which further aggregates local context and yields promising results. But how to effectively apply it to 3D object tracking remains unexplored.

\section{P2B: A Novel Network on Point Set for 3D Object Tracking}
\begin{table}
	\scriptsize
	\begin{center}
		\begin{tabular}{lp{0.6\columnwidth}}
			\toprule
			\textbf{Symbol} &\textbf{Definition} \\
			\hline
			$P_{\rm tmp},P_{\rm sea}$ & Point sets for template and search area. \\
			$q_i,Q$ & Template seed and seeds set. \\
			$r_j,R$ & Search area seed and seeds set\\
			$c_j,C$ & Potential target center and centers set.\\
			$f^{\rm t},F^{\rm t}$ & Target-specific feature and features set\\
			$s^{\rm s}$ & Seed-wise targetness score. \\
			$s^{\rm p}$ & Proposal-wise targetness score.\\
			$p^{\rm t}$ & 3D target proposal. \\
			MLP & Multi-layer perceptron with fully-connected layer, batch normalization and ReLU. \\
			Maxpool & The pooling layer using MAX operation.\\
			\bottomrule
		\end{tabular}
	\end{center}
	\caption{\textbf{Symbols within P2B.}}
	\label{tab:symbol}
\end{table}
\begin{figure}
	\centering	
	\includegraphics[width=8.3cm]{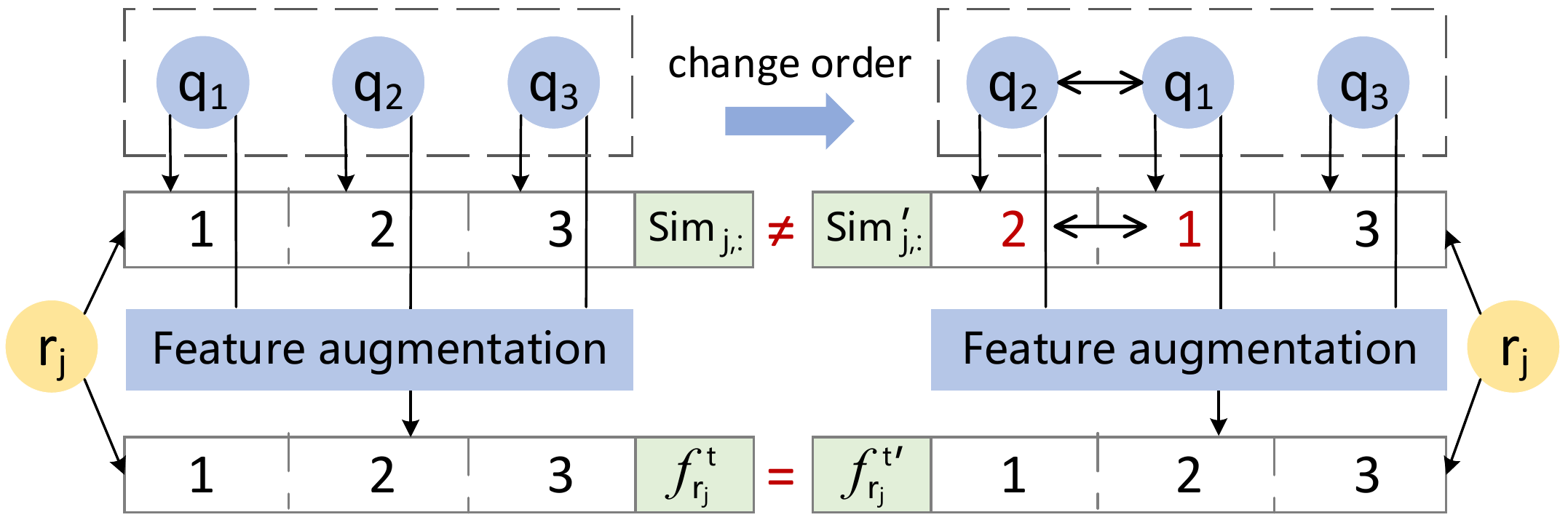}
	\caption{\textbf{The idea of permutation-invariance}. To represent $r_j$, we first compute point-wise similarity $Sim_{j,:}$ between $r_j$ and all template seeds $Q=\{q_i\}^3_{i=1}$. However, $Sim_{j,:}$ keeps changing due to $Q$'s disorder ($Q$'s order can change irregularly). This motivates our feature augmentation for consistent (\textit{i.e.}, permutation-invariant) $f^{\rm t}_{r_j}$. ``$1,2,3$" denote dimensions in $Sim_{j,:}$ and $f^{\rm t}_{r_j}$.}
	\label{permutation-invariance}
	\vspace{-.2cm}
\end{figure}

\subsection{Overview}
In 3D object tracking, we focus on localizing the target (defined by template) in search area frame by frame. We aim to embed template's target clue into search area to predict potential target centers, and execute joint target proposal and verification in an end-to-end manner. P2B has two main parts (Fig.~\ref{fig:P2Bpipline}): 1) target-specific feature augmentation, and 2) 3D target proposal and verification. We first feed template and search area respectively into backbone and obtain their seeds. Then the template seeds help augment the search area seeds with target-specific features. After that, these augmented search area seeds are projected to potential target centers via Hough voting. Seed-wise targetness scores are also calculated to regularize feature learning and strengthen \RM{the discriminative power of these potential target centers}. Then each potential target center clusters its neighbors for 3D target proposal. Proposal with the maximal proposal-wise targetness score is verified as the final result. We will detail them as follows. Main symbols within P2B are defined in Table~\ref{tab:symbol}. For easy comprehension, we also sketch the detailed technical flow in Algorithm~\ref{algorithm}.
\subsection{Target-specific feature augmentation}
\label{section tsfa}

\begin{algorithm}[t]
	\caption{\scriptsize{\textbf{The work flow of P2B}. 
			\protect\\$\Phi$ and $\Theta$ denotes MLP-Maxpool-MLP network operating on feature channel.}}
	\scriptsize
	\label{algorithm}	
	\begin{algorithmic}[1] 
		\Require 
		Points in template ($P_{\rm tmp}$ of size $N_1$) and search area ($P_{\rm sea}$ of size $N_2$).
		\Ensure
		The proposal with the highest $s^{\rm p}$.
		\State 
		\textbf{Feature extraction.} Feed $P_{\rm tmp}$ and $P_{\rm sea}$ into a backbone and respectively get seeds ${Q=\left\{q_i\right\}^{M_1}_{i=1}}$ and ${R=\left\{r_j\right\}^{M_2}_{j=1}}$, with features $f\in\mathbb{R}^{d_1}$. Each seed is represented with its 3D position and $f$ to yield dimension of ${3+d_1}$.
		\State
		\textbf{Point-wise similarity.} Compute point-wise similarity $Sim_{j,:}$ between each seed $r_j$ and $Q$. For all search seeds, we obtain $Sim\in\mathbb{R}^{M_2\times M_1}$.
		\State
		\textbf{Feature augmentation.} Augment each $Sim_{j,:}$ with $Q$ to be of size $M_1\times(1+3+d_1)$. Feed the result into $\Phi$ to get $r_j$'s target-specific feature $f^{\rm t}_ {r_j}\in\mathbb{R}^{d_2}$. $r_j$ is represented with its 3D position and $f^{\rm t}_{r_j}$ to yield dimension of ${3+d_2}$.
		\State
		\textbf{Generating potential target centers.} Each seed $r_j$ 1) predicts a potential target center $c_j$ with feature $f_{c_j}\in\mathbb{R}^{d_2}$ via Hough voting, and 2) is evaluated with seed-wise targetness score $s^{\rm s}_j\in\mathbb{R}$. $c_j$ is represented by concatenating $s^{\rm s}_j$, its 3D position and $f_{c_j}$ to yield dimension of $1+3+d_2$.
		\State
		\textbf{Clustering.} Sample a subset in $C$ to be of size $K$. Generate cluster $T_j$ with ball query for each sampled $c_j$, where $T_j$ contains $n_j$ potential target centers. 
		\State
		\textbf{3D target proposal.} Feed each $T_j$ into $\Theta$ to generate one 3D target proposal $p^{\rm t}_j$ with proposal-wise targetness score $s^{\rm p}_j$. Totally $K$ proposals are predicted.
		\vspace{-0.1cm}	
	\end{algorithmic}
\end{algorithm}
\vspace{-0.2cm}

\begin{figure*}[t]
	\centering
	\includegraphics[width=12cm]{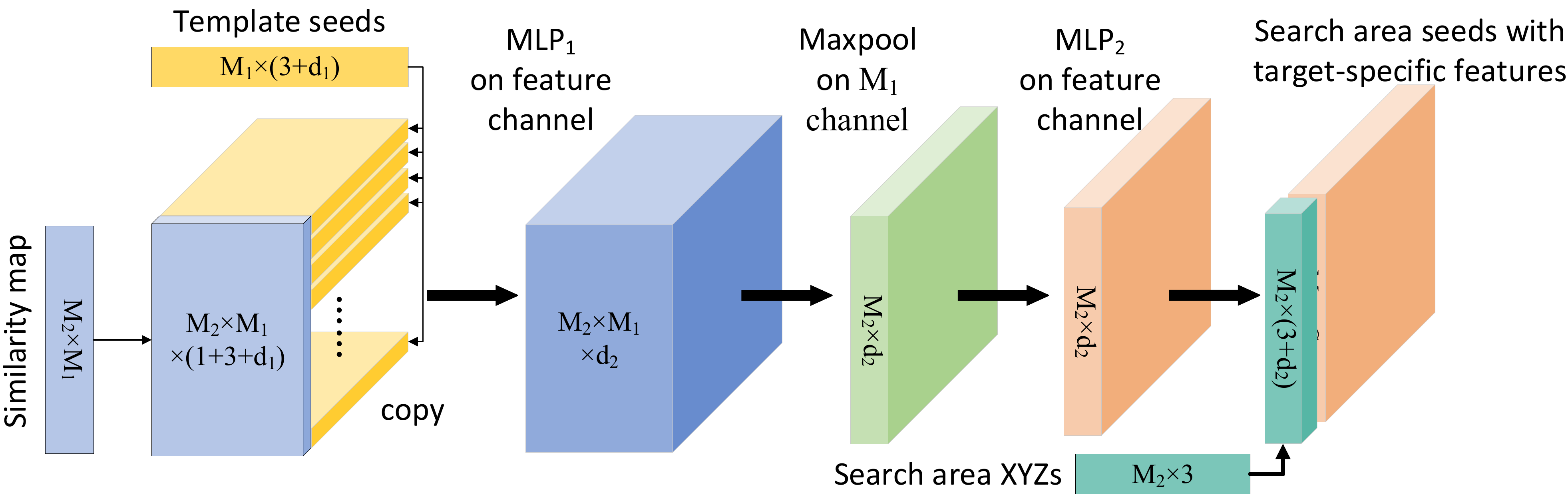}
	\caption{\textbf{Illustration of target-specific feature augmentation}. Our method embeds template's target information into search area seeds while satisfying permutation-invariance.}
	\label{fig:activation}
\end{figure*}

Here we aim to merge template's target information into search area seed to include both global target clue and local tracking clue. We first feed template and search area respectively into feature backbone and obtain their seeds. With the embedded target information in template, we then augment the search area seeds with target-specific features in spirit of pattern matching, which also satisfies permutation-invariance to address point cloud's disorder.

\textbf{Feature encoding on point cloud.}
We feed the points in template $P_{\rm tmp}$ (of size $N_1$) and search area $P_{\rm sea}$ (of size $N_2$) to a feature backbone and obtain ${M_1}$ template seeds ${Q=\left\{q_i\right\}^{M_1}_{i=1}}$ and ${M_2}$ search area seeds ${R=\left\{r_j\right\}^{M_2}_{j=1}}$ with features $f\in\mathbb{R}^{d_1}$.
We applied hierarchical feature learning architecture of PointNet++~\cite{pointnet++} as backbone (but not restricted to it), so that $Q$ and $R$ could preserve local context within $P_{\rm tmp}$ and $P_{\rm sea}$.
Each seed is finally represented with $[x;~f]\in\mathbb{R}^{3+d_1}$ ($x$ denotes the seed's 3D position).

\textbf{Permutation-invariant target-specific feature augmentation}.
To embed $Q$'s target information into $R$, a natural idea is to compute point-wise similarity $Sim$ (of size ${M_2}\times{M_1}$) between $Q$ and $R$, \textit{e.g.}, using cosine distance:
\begin{align}
\label{sim}
{Sim}_{j,i}=\frac{f_{q_i}^\mathrm{T}\cdot f_{r_j}}{\|f_{q_i}\|_2\cdot\|f_{r_j}\|_2}, \forall q_i\in Q,r_j\in R.
\end{align}
Note that $Sim_{j,:}$ (row $j$ in $Sim$) denotes similarity between $r_j$ and all seeds in $Q$.
We may first consider $Sim_{j,:}$ as $r_j$'s target-specific feature.
However, as in Fig. \ref{permutation-invariance}, $Sim_{j,:}$ keeps unstable due to $Q$'s disorder. This contradicts our need for a consistent feature, \textit{i.e.}, a feature invariant to $Q$'s inside permutation.
We accordingly apply symmetric functions (specifically, Maxpool) to ensure permutation-invariance.
As in Fig. \ref{fig:activation}, we first augment each $Sim_{j,:}$ (local tracking clue) with $Q$' spatial coordinates and features (global target clue), yielding a tensor of size $M_1\times{(1+3+d_1)}$.
Then we feed the tensor into network $\Phi$ (MLP-Maxpool-MLP) and obtain $r_j$'s \textit{target-specific feature}, $f^{\rm t}_{r_j}\in\mathbb{R}^{d_2}$.
$r_j$ is finally represented with $[x_{r_j};~f^{\rm t}_{r_j}]\in\mathbb{R}^{3+d_2}$ ($x_{r_j}$ denotes $r_j$'s 3D position).

There are other selections to extract $f^{\rm t}$: leaving out $Q$'s feature, leaving out $Sim$ or adding $R$'s feature. All of them turns inferior in Sec. \ref{exp tsfa}.


\subsection{Target proposal based on potential target \\centers}
\label{section potential target center}
Embedded with target clue, each $r_j$ can directly predict one target proposal. But our intuition is that, individual seed can only capture limited local clue, which may not suffice the final prediction.
We follow the idea within VoteNet~\cite{votenet} to 1) regress the search area seeds into potential target centers via Hough voting, and 2) cluster neighboring centers to leverage the ensemble power and obtain target proposals.

\textbf{Potential target center generation.}
Each seed $r_j$ with feature $f^{\rm t}_{r_j}$ can roughly predict a potential target center $c_j$ via Hough voting. Following VoteNet~\cite{votenet}, the voting module applies MLP to predict the coordinate offset $\Delta{x_j}$ between $r_j$ and ground-truth target center and the residual $\Delta f^{\rm t}_{r_j}$ for $f^{\rm t}_{r_j}$. Hence, $c_j$ is represented with $[x_{c_j};~f_{c_j}]\in\mathbb{R}^{3+d_2}$ where $x_{c_j}=x_{r_j}+\Delta{x_{r_j}}$ and $f_{c_j}=f^{\rm t}_{r_j}+\Delta f^{\rm t}_{r_j}$.
The loss for $\Delta x_j$ is termed as
\begin{align}
\label{loss_anchor}
{L}_{\rm reg}=\frac{1}{M_{\rm ts}}\sum_j\left\|\Delta{x_j}-\Delta{gt_j}\right\|\cdot\mathbb{I}[r_j \mathrm{~on~target}].
\end{align}
Here, $\Delta{gt_j}$ denotes the ground-truth offset from $r_j$ to the target center; $\mathbb{I}(\cdot)$ indicates that we only train those seeds located on the surface of ground-truth target; $M_{\rm ts}$ denotes the number of trained seeds.


\textbf{Clustering and Target proposal.}
For each $c_j$, we use ball query \cite{pointnet++} to generate cluster $T^{\rm t}_j$ with radius $R$: $T^{\rm t}_j=\{c_k|\left\|c_k-c_j\right\|_2<R\}$.
Since neighboring clusters may capture similar region-level context, we sample a subset of size $K$ in all potential target centers as cluster centroids for efficiency.
In Sec. \ref{exp:PropNum}, P2B turns robust to a wide range of $K$s.
Finally we feed each $T^{\rm t}_j$ into $\Theta$ (MLP-Maxpool-MLP) and obtain target proposal $p^{\rm t}_j$ with proposal-wise targetness score $s^{\rm p}_j$ (totally $K$ proposals are generated):
\begin{align}
\label{tgt_prop}
\{p^{\rm t}_j~,s^{\rm p}_j\}=\Theta(T^{\rm t}_j).
\end{align}
$p^{\rm t}_j$ has $4$ parameters: offsets for 3D position and rotation in X-Y plane. We will detail how to learn $\Theta$ in Sec. \ref{section final prediction}.

\subsection{Improved target proposal with seed-wise targetness score}
\label{section target verification}
We consider each seed with target-specific feature can be directly assessed with its targetness to 1) regularize earlier feature learning and 2) strengthen the representation of its predicting potential target center. Therefore, we can obtain target proposals with higher quality.

\textbf{Seed-wise targetness score $s^{\rm s}$.}
We learn a MLP to generate $s^{\rm s}_j$ for each $r_j$. Those search area seeds located on the surface of ground-truth target are regarded as positives, and the extra as negatives. We use a standard binary cross entropy loss ${L}_{\rm cla}$ for $s^{\rm s}$. Since $s^{\rm s}_j$ tightly relates to $f^{\rm t}_{r_j}$, ${L}_{\rm cla}$ can explicitly constrain the point feature learning and consequent target-specific feature augmentation.

\textbf{Improved target proposal.}
Inheriting more discriminative power from $s^{\rm s}_j$, we update $c_j$'s representation with $[s^{\rm s}_{j};~x_{c_j};~f_{c_j}]\in\mathbb{R}^{1+3+d_2}$. Sequentially, we update clusters with ball query and target proposals with Equation (\ref{tgt_prop}). We consider that, $s^{\rm s}$ can implicitly help pick out representative potential target centers to benefit final target proposal.

\subsection{Final target verification}
\label{section final prediction}

With $K$ proposals generated from above (refer to $\Theta$ in Equation (\ref{tgt_prop})), proposal with the highest proposal-wise targetness score is verified as the final tracking result.

We follow VoteNet~\cite{votenet} to learn $\Theta$. Specifically, we consider proposals whose centers near the target center (within 0.3 meters) as positives and those faraway (by more than 0.6 meters) as negatives. Other proposals are left unpenalized. We use a standard binary cross entropy loss ${L}_{\rm prop}$ for $s^{\rm p}_j$. As for $p^{\rm t}_j$, only the positives' box parameters are supervised via Huber (smooth-L1 \cite{fasterRCNN}) loss ${L_{\rm box}}$.
We aggregate all the mentioned losses as our final loss $L$:
\begin{align}
\label{loss_final}
{L}={L}_{\rm reg}+\gamma_1{L_{\rm cla}}+\gamma_2{L_{\rm prop}}+\gamma_3{L_{\rm box}}.
\end{align}
Here $\gamma_1(=0.2)$, $\gamma_2(=1.5)$ and $\gamma_3(=0.2)$ are used to normalize all the component losses to be of the same scale.

\section{Experiments}

We applied KITTI tracking dataset \cite{KITTI} (with point clouds scanned using lidar) as benchmark. We followed settings in \cite{leveraging} (shortened as SC3D by us for simplicity) in data split, tracklet generation\footnote{Frames containing the same target instance, \textit{e.g.}, a car, are concatenated by time order to form a tracklet. } and evaluation metric for fair comparisons. Since cars in KITTI appear in largest quantity and diversity, we mainly focused on car tracking and perform ablation study on it as in SC3D. We also did extensive experiments with other three target types (Pedestrain, Van, Cyclist) for better comparisons.

\subsection{Experimental setting}

\subsubsection{Dataset}

~~~~Since ground truth for test set in KITTI is inaccessible offline, we used its training set to train and test our P2B. This tailored dataset had 21 outdoor scenes and 8 types of targets. We generated tracklets for target instances within all videos and split the dataset as follows: scenes 0-16 for training, 17-18 for validation, and 19-20 for testing.

\begin{figure}[t]
	\centering
	\includegraphics[width=7cm]{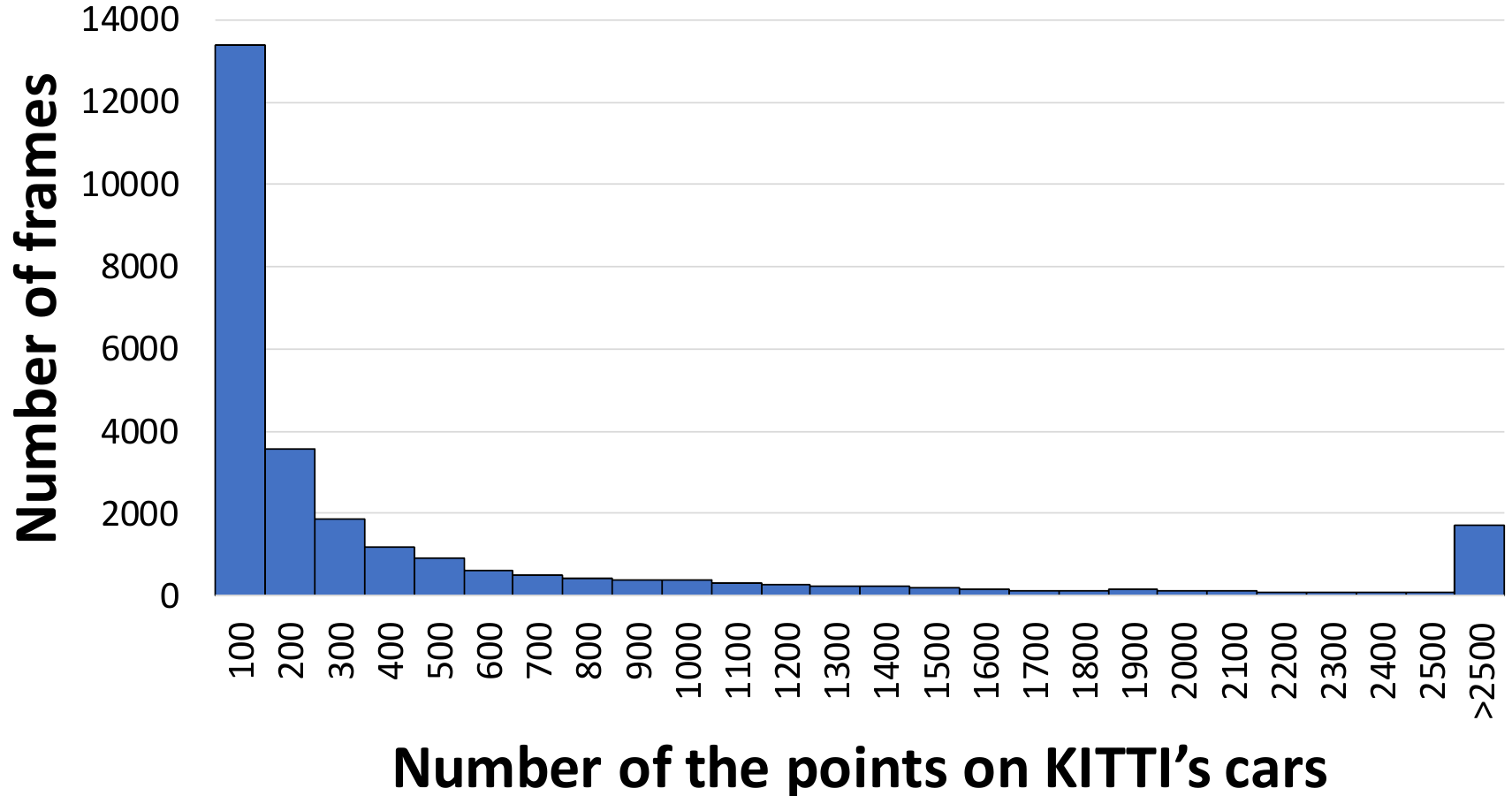}
	\caption{\textbf{Histogram for number of points on KITTI's cars} to exemplify the sparsity of points on target. }
	\label{fig:data_sta}
\end{figure}

\textbf{Point cloud's sparsity.} Though each frame reports an average of 120k points, we suppose the points on target might be quite sparse with general occlusion and lidar's defect on distant objects. To validate our idea, we counted the number of points on KITTI's cars in Fig.~\ref{fig:data_sta}. We can observe that about 34\% cars held fewer than 50 points. The situation may be worse on smaller-size pedestrians and cyclists. This sparsity imposes great challenge onto point cloud based 3D object tracking.

\subsubsection{Evaluation metric}

~~~~We used One Pass Evaluation (OPE) \cite{OPE} to measure Success and Precision of different methods. ``Success" is defined as IOU between predicted box and ground-truth (GT) box. ``Precision" is defined as AUC for errors (distance between two boxes' centers) from 0 to 2m.

\subsubsection{Implementation details}
\label{Sec:Implementation details}

%

\textbf{~~~~Template and search area.} For template\footnote{Template and search area are in forms of point clouds. GT and result are in forms of 3D boxes.}, we collected and normalized its points to $N_1=512$ ones with randomly abandoning or duplicating. For search area, we similarly collected and normalized the points to $N_2=1024$ ones. The ways to generate template and search area differ in training and testing as detailed below.

\textbf{Network architecture.} We adopted PointNet++~\cite{pointnet++} as our backbone. We tailored it to contain three set-abstraction (SA) layers, with receptive radius of $0.3,~0.5,~0.7$ meters, and 3 times of half-size down-sampling. This yielded $M_1=64(=N_1/2^3)$ template seeds and $M_2=128(=N_2/2^3)$ search area seeds. We applied random sampling, and removed up-sampling layers in PointNet++ due to points' sparsity. The output feature was of $d_1=256$ dimensions.

Throughout our method, all used MLPs had three layers. The size of these layers was $256$ (hence $d_2=256$) except that of the last layers ($\mathrm{size_{ly}}$) in following MLPs:
\begin{itemize}
	\setlength{\itemsep}{0pt}
	\setlength{\parsep}{0pt}
	\setlength{\parskip}{0pt}	
	\item For MLP to predict $s^{\rm s}$, $\mathrm{size_{ly}}=1$.
	\item For $\Theta$ to predict $s^{\rm p}$ and $p^{\rm t}$, $\mathrm{size_{ly}}=5$.
\end{itemize}

\textbf{Clustering}. $K=64$ randomly sampled potential target centers clustered the neighbors within $R=0.3$ meters.

\textbf{Training.}
1) Data Augmentation: we applied random offset on previous GT and fused point clouds within the result box and the first GT for more template samples; we enlarged the current GT by 2 meters to include background (negative seeds), applied similar random offset and collected inside point cloud for more search area samples.
2) We trained P2B from scratch with the augmented samples. We applied Adam optimizer \cite{Adam}. Learning rate was initially 0.001 and decreased by 5 times after 10 epochs. Batch size was 32. In practice, we observed P2B converged to a satisfying result after about 40 epochs.

\textbf{Testing.} We used the trained P2B to infer 3D bounding boxes within tracklets frame by frame. For the current frame, template initially adopted the first GT's point cloud and then fusion of the first GT's and previous result's point clouds. We enlarged previous result by 2 meters in current frame and collected inside point cloud to obtain search area.

\begin{table}[]
	\scriptsize
	\begin{center}
		\begin{tabular}{ccccc}   \toprule
			\multicolumn{1}{c}{} & Method     & Previous result  & Previous GT    & Current GT \\  \hline
			\multirow{2}{*}{Success}  & SC3D~\cite{leveraging}       & 41.3             & 64.6           & 76.9      \\  
			& P2B~(ours)  & \textbf{56.2}    & \textbf{82.4}  & \textbf{84.0}    \\ \hline 
			\multirow{2}{*}{Precision} & SC3D~\cite{leveraging}       & 57.9             & 74.5           & 81.3      \\   
			& P2B~(ours)  & \textbf{72.8}    & \textbf{90.1}  & \textbf{90.3}   \\ \hline 
		\end{tabular}
	\end{center}
	\caption{\textbf{Comprehensive comparison with SC3D.} The right three columns differ in their ways to generate search area.}
	\label{tab:backbone}
\end{table}

\begin{table}
	\scriptsize
	\begin{center}
		\begin{tabular}{@{\hspace{0.1pt}}p{0.07\columnwidth}<{\centering}ccccccc@{\hspace{0.1pt}}} \toprule
			\multirow{2}{*}{} & Method     & Car  & Pedestrian    & Van & Cyclist  & Mean \\  
			\multicolumn{1}{c}{} & Frame Number     & 6424  & 6088    & 1248 & 308  & 14068 \\  \hline
			\multirow{2}{*}{Success}  & SC3D~\cite{leveraging}       & 41.3             & 18.2           & 40.4   & \textbf{41.5}  & 31.2   \\  
			& P2B~(ours)  & \textbf{56.2}    & \textbf{28.7}  & \textbf{40.8}   & 32.1  & \textbf{42.4} \\ \hline 
			\multirow{2}{*}{Precision} & SC3D~\cite{leveraging}       & 57.9             & 37.8           & 47.0   & \textbf{70.4}  & 48.5   \\ 
			& P2B~(ours)  & \textbf{72.8}    & \textbf{49.6}  & \textbf{48.4}  & 44.7  & \textbf{60.0} \\ \hline 
		\end{tabular}
	\end{center}
	\caption{\textbf{Extensive comparisons with SC3D.} The right five colu-mns show results with different target types and their Mean.}
	\label{tab:category}
\end{table}
\subsection{Comprehensive comparisons}

We only compared our P2B with SC3D~\cite{leveraging}, the first and only work on point cloud based 3D object tracking.
We reported results for 3D car tracking in Table \ref{tab:backbone}.

We generated search area centered on previous result, previous GT or current GT.
Using previous result as the search center meets the requirement of real scenarios, while using previous GT helps approximately assess short-term tracking performance.
For the two situations, SC3D applies Kalman filtering to generate proposals.
Using current GT is unreasonable, but is considered in SC3D to approximate exhaustive search and assess SC3D's discriminative power.
Specifically, SC3D conducts grid search around target center to include GT box in generated proposals.
However, P2B clusters potential target centers to generate proposals without explicit dependence on GT box. \textit{I.e.}, P2B may adapt to various scenarios while SC3D could degrade when the GT boxes are removed as demonstrated in Table \ref{tab:backbone} .
Comprehensively, P2B outperformed SC3D by a large margin.
All later experiments adopted the more realistic setting of using previous result (``Testing" in Sec. \ref{Sec:Implementation details}).

\textbf{Extensive comparisons.} We further compared P2B with SC3D on Pedestrian, Van, and Cyclist (Table \ref{tab:category}). P2B outperformed SC3D by $\sim$10\% on average.
P2B's advantage turned significant on data-rich Car and Pedestrian.
But P2B degraded when training data decreased as was the case for Van and Cyclist.
We conjecture that P2B may rely on more data to learn better networks especially when regressing potential target centers. Comparatively, SC3D needs relatively less data to suffice similarity measuring between two regions.
To validate this, we used the model trained on data-rich Car to test Van, with the belief that car resembles van and contains potentially transferable information. As expected, the Success/Precision result of P2B showed an improved 49.9/59.9 (original: 40.8/48.4), while SC3D reported a declined 37.2/45.9 (original: 40.4/47.0).

\begin{table}
	\scriptsize
	\begin{center}
		\begin{tabular}{cccc}
			\toprule
			\textbf{Ways for \textit{tsfa}} & Success & Precision \\
			\hline
			Our default setting & 56.2 & \textbf{72.8}  \\
			Without template features & 55.6 & 70.9 \\
			Without similarity map & 52.7 & 69.4 \\
			With search area features A & \textbf{56.8} & 72.6 \\
			With search area features B & 49.3 & 64.8 \\
			\hline
		\end{tabular}
	\end{center}
	\caption{\textbf{Different ways for target-specific feature augmentation (\textit{tsfa})}. Methods for obtaining search features A and B are illustrated in Fig. \ref{fig:add_features}.}
	\label{tab:tsfa}
\end{table}

\begin{figure}[t]
	\centering
	\includegraphics[width=7cm]{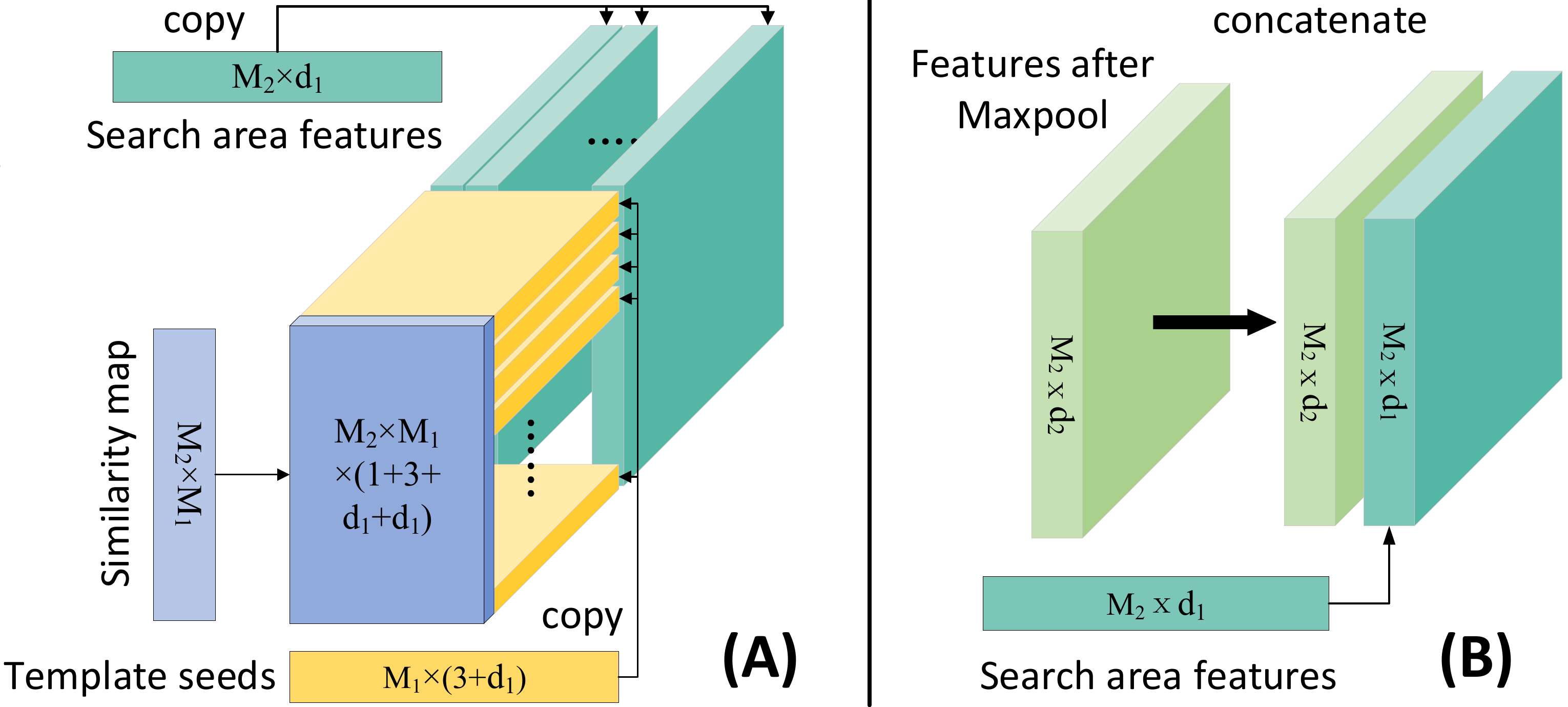}
	\caption{\textbf{Two ways to include search area features in target-specific feature augmentation}. For A we duplicated search area seeds' features and attached them after template features' duplications along each column of similarity map; for B we concatenated the search area feature with the feature after Maxpool (Fig. \ref{fig:activation}).}
	\label{fig:add_features}
\end{figure}


\subsection{Ablation study}
\subsubsection{Ways for target-specific feature augmentation}
\label{exp tsfa}

~~~~Besides our default setting in P2B (Sec. \ref{section tsfa}), there are another four possible ways for feature augmentation: removing (the duplication of) template features, removing the similarity map, using search area feature A and B (Fig. \ref{fig:add_features}).

We compared the five settings in Table \ref{tab:tsfa}. Here removing template features or similarity map degraded by about 1\% or 3\%, which validates the contributions of these two parts in our default setting. Search area feature A and B did not improve or even harm the performance. Note that we already combined template features in both conditions. This may reveal that search area features only capture spatial context rather than target clue, and hence turns useless for target-specific feature augmentation. In comparison, our default setting brings with richer target clue from template seeds to yield a more ``directed" proposal generation.

\begin{figure*}[t]
	\centering
	\includegraphics[width=15.5cm]{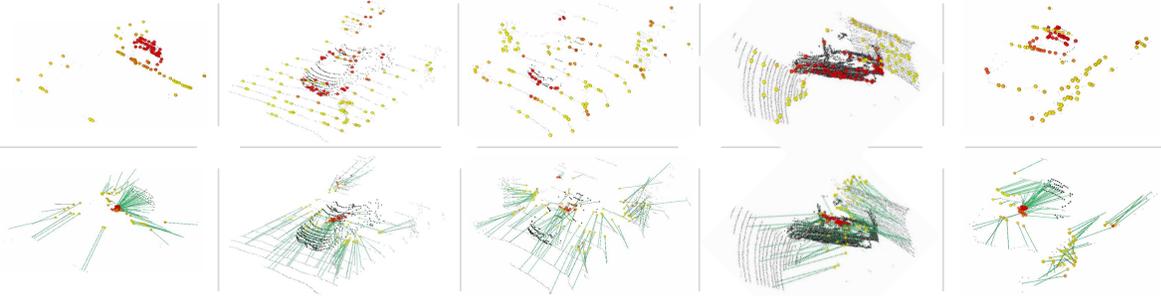}
	\caption{\textbf{Illustration of seed-wise targetness scores and potential target centers.} Green lines show projection from seeds (colored points in the first row) to potential target centers (colored points in the second row). We marked those informative points, \textit{i.e.}, with higher targetness scores, in red and opposite in yellow. Paired seed and potential center were marked in the same color to show correlation.}
	\label{fig:qualitative}
\end{figure*}
\subsubsection{Effectiveness of seed-wise targetness}

~~~~In Sec. \ref{section target verification}, we obtain seed-wise targetness scores $s^{\rm s}$ and concatenate them with potential target centers to guide the proposal and verification. Here we tested P2B without this concatenation or even the whole branch of $s^{\rm s}$ (Table \ref{tab:target verification}). We can observe that leaving out concatenation dropped the performance by $\sim$1\%, while removing the whole branch dropped by $\sim$3\%. This verifies that $s^{\rm s}$ offers good supervision on learning the whole network for improved target proposal and verification.

\subsubsection{Robustness with different number of proposals}
\label{exp:PropNum}

\begin{figure}[]
	\centering
	\includegraphics[scale=0.48]{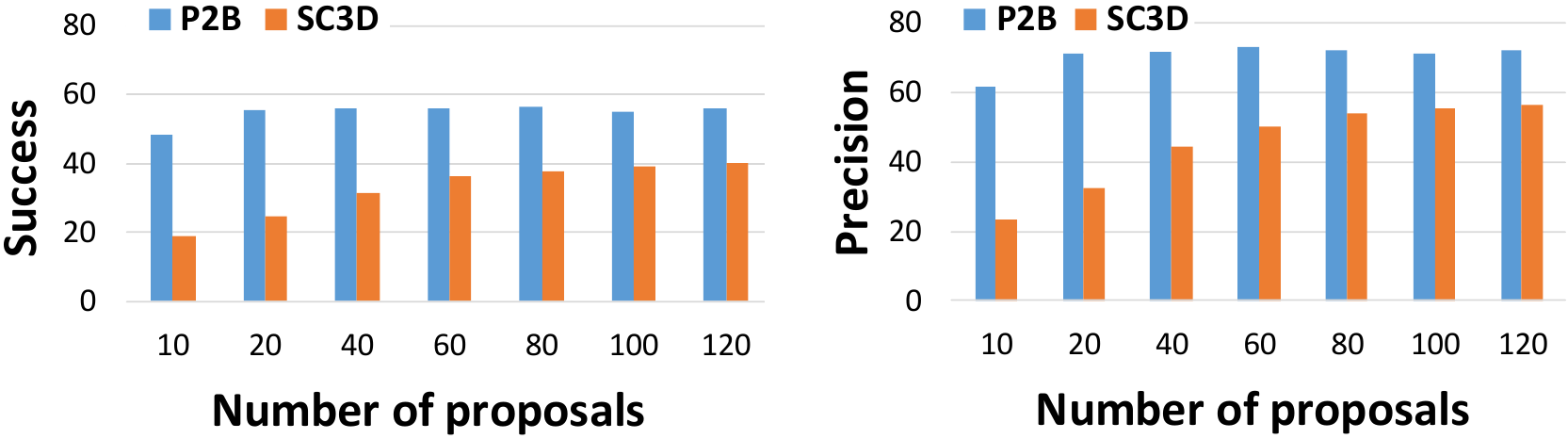}
	\caption{\textbf{Different number of the proposals} to show our method is compatible with a wide range of parameters.}
	\label{fig:PropNum}
	\vspace{-0.08cm}
\end{figure}

~~~~We tested P2B (without re-training) and SC3D with different number of proposals. From the results in Fig. \ref{fig:PropNum}, P2B obtained satisfying results even with only 20 proposals. But SC3D degraded dramatically when using less than 40 proposals. To conclude, P2B turns more robust to less number of proposals, showing that P2B can generate proposals with both higher quality and efficiency.
\begin{table}
	\scriptsize
	\begin{center}
		\begin{tabular}{cccc}
			\toprule
			\textbf{Ways for using $s^{\rm s}$} & Success & Precision \\
			\hline
			Our default setting & \textbf{56.2} & \textbf{72.8}  \\
			Without concatenation & 55.1 & 70.8 \\
			Without the whole branch of $s^{\rm s}$ & 52.6 & 67.4 \\
			\hline
		\end{tabular}
	\end{center}
	\caption{\textbf{Effectiveness of seed-wise targetness}.}
	\label{tab:target verification}
\end{table}
\begin{figure*}[t]
	\centering
	\includegraphics[width=14cm]{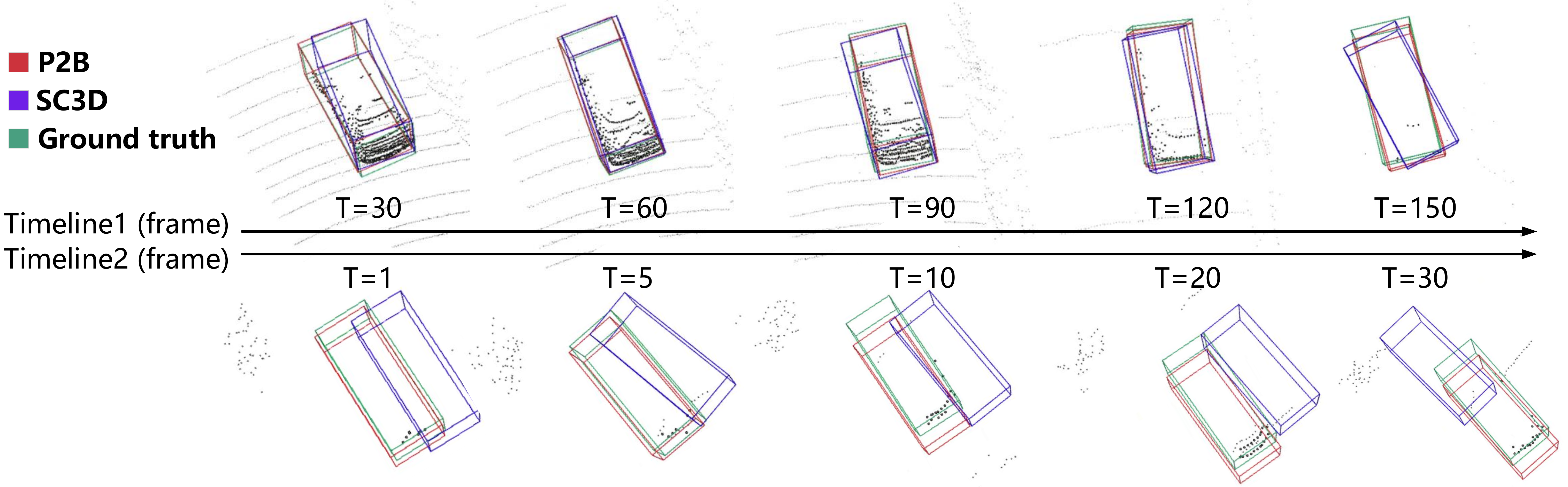}
	\caption{\textbf{Advantageous cases of our P2B compared with SC3D.} We can observe P2B's advantage over SC3D in both dense (the first-row sequence) and sparse (the second-row sequence) scenarios, especially for the latter.}
	\label{fig:qualitative1}
\end{figure*}
\subsubsection{Ways for template generation}

~~~~For template generation, SC3D concatenates the points in all previous results while P2B concatenates the points within the first GT and previous result to update template for efficiency.
Here we reported results with four settings for template generation: the first GT, the previous result, the fusion of the first GT and previous result, and all previous results.
Results in Table \ref{tab:shape_aggregation} show P2B's consistent advantage over SC3D in all settings, even in ``All previous shapes" where P2B reported degraded result. We attribute the degradation to that 1) we did not include shape completion~\cite{leveraging} and 2) we did not train P2B with all previous results while SC3D considered both.

\begin{figure*}[!tp]
	\centering
	\includegraphics[scale=0.19]{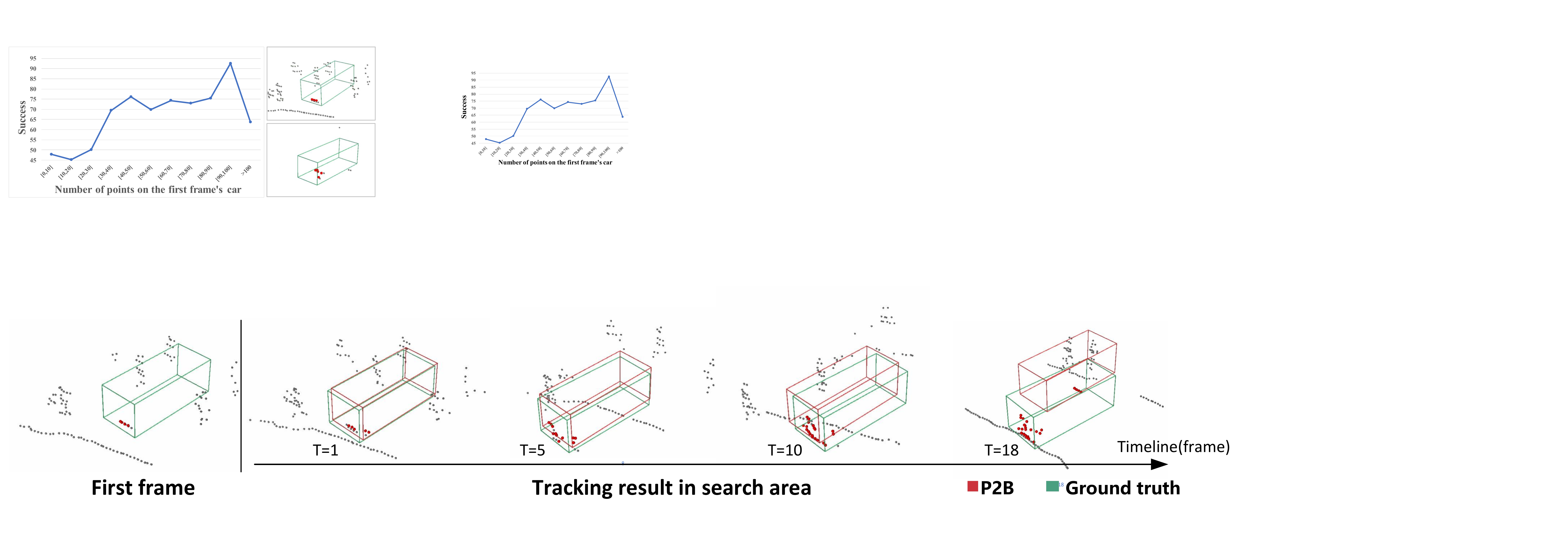}
	\caption{\textbf{Failure cases of P2B when the initial template contained few informative points}. }
	\label{fig:failure}
\end{figure*}

\begin{table}
	\scriptsize
	\begin{center}
		\centering
		\begin{tabular}{@{\hspace{0.05pt}}p{0.25\columnwidth}<{\centering}p{0.13\columnwidth}<{\centering}p{0.1\columnwidth}<{\centering}p{0.01cm}p{0.13\columnwidth}<{\centering}p{0.13\columnwidth}<{\centering}@{\hspace{0.05pt}}}
			\toprule
			\textbf{Source of} & \multicolumn{2}{c}{Success } && \multicolumn{2}{c}{Precision } \\
			\cline{2-3}\cline{5-6}
			\textbf{template points} & P2B (ours)  & SC3D~\cite{leveraging}  && P2B (ours) & SC3D~\cite{leveraging} \\
			\hline
			The First GT  & \textbf{46.7 }  & 31.6  && \textbf{59.7}   & 44.4 \\
			Previous result  &\textbf{53.1}  & 25.7  && \textbf{68.9}   & 35.1 \\
			{First \& Previous}  & \textbf{56.2}   & 34.9  && \textbf{72.8 }  & 49.8 \\
			All previous results  & \textbf{51.4}   & 41.3 && \textbf{66.8 }  & 57.9 \\
			\hline
		\end{tabular}
	\end{center}
	\caption{\textbf{Different ways for template generation}. ``First \& Previous" denotes ``The first GT and Previous result". }
	\label{tab:shape_aggregation}
\end{table}

\subsection{Qualitative analysis}
\subsubsection{Advantageous cases}

~~~~We first exemplified our target-specific feature's discriminative power in Fig. \ref{fig:qualitative}.
The first row visualizes seeds' targetness scores to demonstrate their possibility of belonging to the target (Car).
We can observe that P2B had learnt to discriminate the target seeds from the background ones.
The second row visualizes how P2B projects seeds to potential target centers.
We can observe that the potential centers with more target information gathered tightly around GT target center, which further validates our discriminative target-specific features. Besides, P2B can address the occlusion because it can generate groups of informative potential target centers for final prediction.

We then visualize P2B's advantage over SC3D to address point cloud's sparsity in Fig. \ref{fig:qualitative1}. We can observe that in the sparse scenarios where SC3D tracked off course or even failed, our predicted box held tight to the target center.

\subsubsection{Failure cases}

~~~~Here we searched for tracklets where P2B failed and found that most failure cases arose when initial template in the first frame was too sparse and hence yielded little target information. As exemplified in Fig. \ref{fig:failure}, when P2B faced such case and tracked off course with cluttered background, points from the initial template cannot modify current erroneous predictions and re-obtain an informative template. This failure may also reveal that P2B inherits target information from template instead of search area.

\begin{figure}[]
	\centering
	\includegraphics[width=7.4cm]{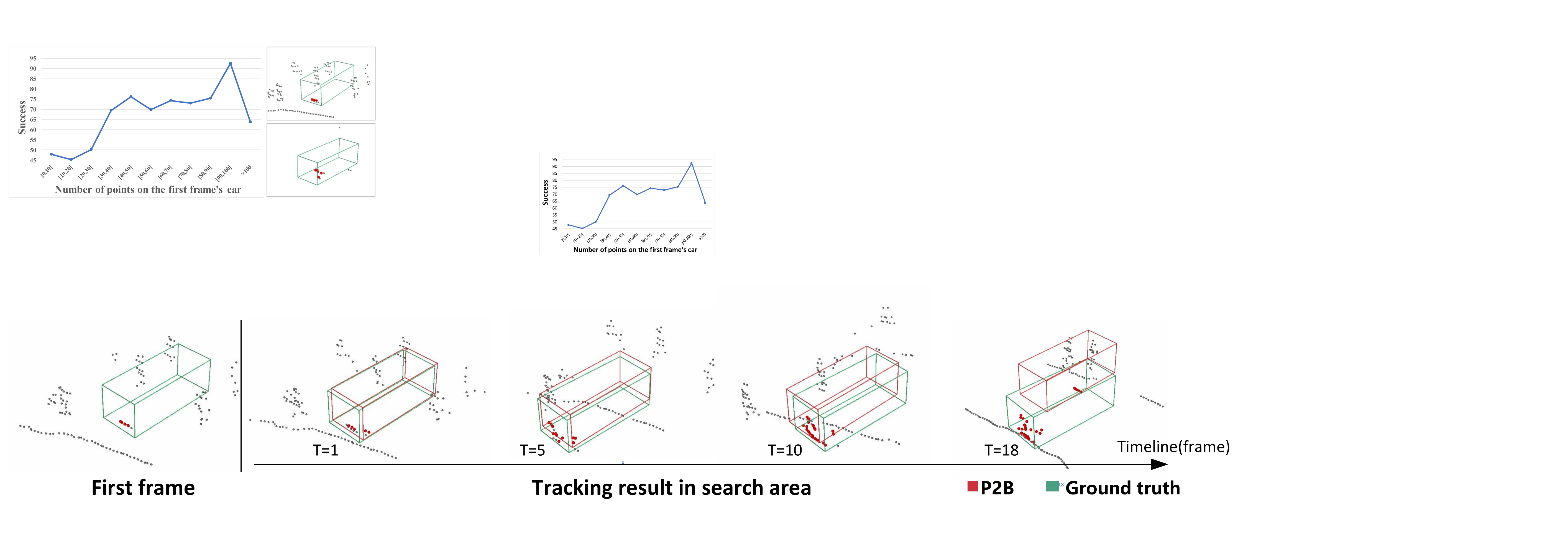}
	\caption{\textbf{The influence of the number of points on the first frame's car to our method}. We counted the average Success for each interval (horizontal axis) in the test set.}
	\label{fig:failure-analysis}
\end{figure}

We believe that when fed with more points containing potentially rich target information, P2B could generate proposals with higher quality to yield better results. Our intuition is validated in Fig. \ref{fig:failure-analysis}.

\subsection{Running speed}

Here we averaged the running time of all test frames for car to measure P2B's speed. P2B achieved 45.5 FPS, including 7.0 ms for processing point cloud, 14.3 ms for network forward propagation and 0.9ms for post-processing, on a single NVIDIA 1080Ti GPU. SC3D in default setting ran with 1.8 FPS on the same platform.

\section{Conclusions}
In this work we propose a novel point-to-box (P2B) network for 3D object tracking. We focus on embedding the target information within template into search space and formulate an end-to-end method for point-driven target proposal and verification jointly. P2B operates on sampled seeds instead of 3D boxes to reduce search space by a large margin. Experiments justify our proposition's superiority.

The experiments also reveal that P2B needs more data to obtain satisfying result. Hence, we could expect a less data-dependent P2B while we could also collect more data to handle the issue under this big-data era.
Besides, we could seek better ways for feature augmentation in search area and test our method on more challenging scenarios.

~\\
\noindent\textbf{Acknowledgements}
This work is jointly supported by the National Natural Science Foundation of China (Grant No. U1913602, 61876211 and 61502187), Equipment Pre-research Field Fund of China (Grant No. 61403120405), National Key Laboratory Open Fund of China (Grant No. 6142113180211), and the Fundamental Research Funds for the Central Universities (Grant No. 2019kfyXKJC024).

{\small
\bibliographystyle{ieee_fullname}
\bibliography{egbib}

\begin{thebibliography}{10}\itemsep=-1pt

\bibitem{asvadi3d}
Alireza Asvadi, Pedro Gir{\~a}o, Paulo Peixoto, and Urbano Nunes.
\newblock 3d object tracking using rgb and lidar data.
\newblock In {\em Proc. IEEE International Conference on Intelligent
  Transportation Systems (ITSC)}, 2016.

\bibitem{houghtransform}
D.~H. Ballard.
\newblock Generalizing the hough transform to detect arbitrary shapes.
\newblock {\em Pattern recognition}, 13(2):111--122, 1981.

\bibitem{siamFC}
Luca Bertinetto, Jack Valmadre, Joao~F Henriques, Andrea Vedaldi, and Philip~HS
  Torr.
\newblock Fully-convolutional siamese networks for object tracking.
\newblock In {\em Proc. European Conference on Computer Vision (ECCV)}, 2016.

\bibitem{Bibi3D}
Adel Bibi, Tinahzu Zhang, and Bernard Ghanem.
\newblock 3d part-based sparse tracker with automatic synchronization and
  registration.
\newblock In {\em Proc. IEEE Conference on Computer Vision and Pattern
  Recognition (CVPR)}, 2016.

\bibitem{pointnet}
R.~Qi Charles, Su Hao, Kaichun Mo, and Leonidas~J. Guibas.
\newblock Pointnet: Deep learning on point sets for 3d classification and
  segmentation.
\newblock In {\em Proc. IEEE Conference on Computer Vision and Pattern
  Recognition (CVPR)}, 2017.

\bibitem{chen2018shpr}
Xinghao Chen, Guijin Wang, Cairong Zhang, Tae-Kyun Kim, and Xiangyang Ji.
\newblock Shpr-net: Deep semantic hand pose regression from point clouds.
\newblock {\em IEEE Access}, pages 43425--43439, 2018.

\bibitem{comport2004robust}
Andrew~I Comport, {\'E}ric Marchand, and Fran{\c{c}}ois Chaumette.
\newblock Robust model-based tracking for robot vision.
\newblock In {\em Proc. IEEE/RSJ International Conference on Intelligent Robots
  and Systems (IROS)}, 2004.

\bibitem{fan2019siamese}
Heng Fan and Haibin Ling.
\newblock Siamese cascaded region proposal networks for real-time visual
  tracking.
\newblock In {\em Proc. IEEE Conference on Computer Vision and Pattern
  Recognition (CVPR)}, 2019.

\bibitem{HandPointNet}
Liuhao Ge, Yujun Cai, Junwu Weng, and Junsong Yuan.
\newblock Hand pointnet: 3d hand pose estimation using point sets.
\newblock In {\em Proc. IEEE Conference on Computer Vision and Pattern
  Recognition (CVPR)}, 2018.

\bibitem{KITTI}
Andreas Geiger, Philip Lenz, and Raquel Urtasun.
\newblock Are we ready for autonomous driving? the kitti vision benchmark
  suite.
\newblock In {\em Proc. IEEE Conference on Computer Vision and Pattern
  Recognition (CVPR)}, 2012.

\bibitem{leveraging}
Silvio Giancola, Jesus Zarzar, and Bernard Ghanem.
\newblock Leveraging shape completion for 3d siamese tracking.
\newblock In {\em Proc. IEEE Conference on Computer Vision and Pattern
  Recognition (CVPR)}, 2019.

\bibitem{gordon2004beyond}
Neil Gordon, B Ristic, and S Arulampalam.
\newblock Beyond the kalman filter: Particle filters for tracking applications.
\newblock {\em Artech House, London}, 2004.

\bibitem{held2016learning}
David Held, Sebastian Thrun, and Silvio Savarese.
\newblock Learning to track at 100 fps with deep regression networks.
\newblock In {\em Proc. European Conference on Computer Vision (ECCV)}, 2016.

\bibitem{huang2019GlobalTrack}
Lianghua Huang, Xin Zhao, and Kaiqi Huang.
\newblock Globaltrack: A simple and strong baseline for long-term tracking.
\newblock {\em arXiv preprint arXiv:1912.08531}, 2019.

\bibitem{kartobject}
Ugur Kart, Alan Lukezic, Matej Kristan, Joni-Kristian Kamarainen, and Jiri
  Matas.
\newblock Object tracking by reconstruction with view-specific discriminative
  correlation filters.
\newblock In {\em Proc. IEEE Conference on Computer Vision and Pattern
  Recognition (CVPR)}, 2019.

\bibitem{RGBDtracker}
Matas~J. Kart~U, Kamarainen J~K.
\newblock How to make an rgbd tracker?
\newblock In {\em Proc. European Conference on Computer Vision (ECCV)}, 2018.

\bibitem{Adam}
Diederik~P. Kingma and Jimmy Ba.
\newblock Adam: {A} method for stochastic optimization.
\newblock In {\em Proc. International Conference on Learning Representations
  (ICLR)}, 2015.

\bibitem{Escape}
Roman Klokov and Victor Lempitsky.
\newblock Escape from cells: Deep kd-networks for the recognition of 3d point
  cloud models.
\newblock In {\em Proc. IEEE International Conference on Computer Vision
  (ICCV)}, 2017.

\bibitem{houghvoting}
Bastian Leibe, Ales Leonardis, and Bernt Schiele.
\newblock Robust object detection with interleaved categorization and
  segmentation.
\newblock {\em International Journal of Computer Vision}, 77(1--3):259--289,
  2008.

\bibitem{li2019siamrpn}
Bo Li, Wei Wu, Qiang Wang, Fangyi Zhang, Junliang Xing, and Junjie Yan.
\newblock Siamrpn++: Evolution of siamese visual tracking with very deep
  networks.
\newblock In {\em Proc. IEEE Conference on Computer Vision and Pattern
  Recognition (CVPR)}, 2019.

\bibitem{siamRPN}
Bo Li, Junjie Yan, Wei Wu, Zheng Zhu, and Xiaolin Hu.
\newblock High performance visual tracking with siamese region proposal
  network.
\newblock In {\em Proc. IEEE Conference on Computer Vision and Pattern
  Recognition (CVPR)}, 2018.

\bibitem{Point_To_Pose}
Shile Li and Dongheui Lee.
\newblock Point-to-pose voting based hand pose estimation using residual
  permutation equivariant layer.
\newblock In {\em Proc. IEEE Conference on Computer Vision and Pattern
  Recognition (CVPR)}, 2019.

\bibitem{li2018pointcnn}
Yangyan Li, Rui Bu, Mingchao Sun, Wei Wu, Xinhan Di, and Baoquan Chen.
\newblock Pointcnn: Convolution on x-transformed points.
\newblock In {\em Proc. Advances in Neural Information Processing Systems
  (NIPS)}, 2018.

\bibitem{context}
Ye Liu, Xiao-Yuan Jing, Jianhui Nie, Hao Gao, Jun Liu, and Guo-Ping Jiang.
\newblock Context-aware three-dimensional mean-shift with occlusion handling
  for robust object tracking in rgb-d videos.
\newblock {\em IEEE Transactions on Multimedia}, pages 664--677, 2018.

\bibitem{luo2018fast}
Wenjie Luo, Bin Yang, and Raquel Urtasun.
\newblock Fast and furious: Real time end-to-end 3d detection, tracking and
  motion forecasting with a single convolutional net.
\newblock In {\em Proc. IEEE Conference on Computer Vision and Pattern
  Recognition (CVPR)}, 2018.

\bibitem{machida2012human}
Eiji Machida, Meifen Cao, Toshiyuki Murao, and Hiroshi Hashimoto.
\newblock Human motion tracking of mobile robot with kinect 3d sensor.
\newblock In {\em Proc. SICE Annual Conference (SICE)}, 2012.

\bibitem{pieropan2015robust}
Alessandro Pieropan, Niklas Bergstr{\"o}m, Masatoshi Ishikawa, and Hedvig
  Kjellstr{\"o}m.
\newblock Robust 3d tracking of unknown objects.
\newblock In {\em Proc. IEEE International Conference on Robotics and
  Automation (ICRA)}, 2015.

\bibitem{votenet}
Charles~R Qi, Or Litany, Kaiming He, and Leonidas~J Guibas.
\newblock Deep hough voting for 3d object detection in point clouds.
\newblock In {\em Proc. IEEE International Conference on Computer Vision
  (ICCV)}, 2019.

\bibitem{frustum}
Charles~R Qi, Wei Liu, Chenxia Wu, Hao Su, and Leonidas~J Guibas.
\newblock Frustum pointnets for 3d object detection from rgb-d data.
\newblock In {\em Proc. IEEE Conference on Computer Vision and Pattern
  Recognition (CVPR)}, 2018.

\bibitem{pointnet++}
Charles~Ruizhongtai Qi, Li Yi, Hao Su, and Leonidas~J Guibas.
\newblock Pointnet++: Deep hierarchical feature learning on point sets in a
  metric space.
\newblock In {\em Proc. Advances in Neural Information Processing Systems
  (NIPS)}, 2017.

\bibitem{fasterRCNN}
Shaoqing Ren, Kaiming He, Ross Girshick, and Jian Sun.
\newblock Faster r-cnn: Towards real-time object detection with region proposal
  networks.
\newblock In {\em Proc. Advances in Neural Information Processing Systems
  (NIPS)}, 2015.

\bibitem{PointRCNN}
Shaoshuai Shi, Xiaogang Wang, and Hongsheng Li.
\newblock Pointrcnn: 3d object proposal generation and detection from point
  cloud.
\newblock In {\em Proc. IEEE Conference on Computer Vision and Pattern
  Recognition (CVPR)}, 2019.

\bibitem{tao2016siamese}
Ran Tao, Efstratios Gavves, and Arnold~WM Smeulders.
\newblock Siamese instance search for tracking.
\newblock In {\em Proc. IEEE Conference on Computer Vision and Pattern
  Recognition (CVPR)}, 2016.

\bibitem{wang2017dcfnet}
Qiang Wang, Jin Gao, Junliang Xing, Mengdan Zhang, and Weiming Hu.
\newblock Dcfnet: Discriminant correlation filters network for visual tracking.
\newblock {\em arXiv preprint arXiv:1704.04057}, 2017.

\bibitem{wang2018learning}
Qiang Wang, Zhu Teng, Junliang Xing, Jin Gao, Weiming Hu, and Stephen Maybank.
\newblock Learning attentions: residual attentional siamese network for high
  performance online visual tracking.
\newblock In {\em Proc. IEEE Conference on Computer Vision and Pattern
  Recognition (CVPR)}, 2018.

\bibitem{wang2019fast}
Qiang Wang, Li Zhang, Luca Bertinetto, Weiming Hu, and Philip~HS Torr.
\newblock Fast online object tracking and segmentation: A unifying approach.
\newblock In {\em Proc. IEEE Conference on Computer Vision and Pattern
  Recognition (CVPR)}, 2019.

\bibitem{wang2019GANTrack}
Xiao Wang, Tao Sun, Rui Yang, and Bin Luo.
\newblock Learning target-aware attention for robust tracking with conditional
  adversarial network.
\newblock In {\em Proc. British Machine Vision Conference (BMVC)}, 2016.

\bibitem{OPE}
Yi Wu, Jongwoo Lim, and Ming-Hsuan Yang.
\newblock Online object tracking: A benchmark.
\newblock In {\em Proc. IEEE Conference on Computer Vision and Pattern
  Recognition (CVPR)}, 2013.

\bibitem{Yang_2019_ICCV_STD}
Zetong Yang, Yanan Sun, Shu Liu, Xiaoyong Shen, and Jiaya Jia.
\newblock Std: Sparse-to-dense 3d object detector for point cloud.
\newblock In {\em Proc. IEEE International Conference on Computer Vision
  (ICCV)}, 2019.

\bibitem{zhang2019deeper}
Zhipeng Zhang and Houwen Peng.
\newblock Deeper and wider siamese networks for real-time visual tracking.
\newblock In {\em Proc. IEEE Conference on Computer Vision and Pattern
  Recognition (CVPR)}, 2019.

\bibitem{Zhu2016BeyondLS}
Gao Zhu, Fatih~Murat Porikli, and Hongdong Li.
\newblock Beyond local search: Tracking objects everywhere with
  instance-specific proposals.
\newblock In {\em Proc. IEEE Conference on Computer Vision and Pattern
  Recognition (CVPR)}, 2016.

\bibitem{zhu2018distractor}
Zheng Zhu, Qiang Wang, Bo Li, Wei Wu, Junjie Yan, and Weiming Hu.
\newblock Distractor-aware siamese networks for visual object tracking.
\newblock In {\em Proc. European Conference on Computer Vision (ECCV)}, 2018.

\end{thebibliography}
}

\end{document}